\documentclass[ sigconf,
            ]{nimeart-arxiv}
\usepackage{listings}
\usepackage{tabularx}
\usepackage{xcolor}
\usepackage{subcaption}
\usepackage{stfloats}
\usepackage{tikz}
\usepackage{float}
\usetikzlibrary{positioning, arrows.meta, fit, calc}
\setcopyright{cc}
\copyrightyear{2026}
\acmYear{2026}
\acmDOI{}

\acmConference[NIME '26]{International Conference on New Interfaces for Musical Expression}{June 23--26, 2026}{London, UK}
\acmISBN{}

\begin{document}

\title{Choreographing the Way of Water: A Computational Framework for Aquatic Robotic Art}

% Author Block Organization
\author{Aswin Ramachandran}
% \authornote{Authors 1--5 contributed equally to this research.}
\email{aramachandra@ethz.ch}
\affiliation{%
  \institution{ETH Zurich}
  \city{Zurich}
  \country{Switzerland}
}

\author{Christopher Golling}
\email{cgolling@student.ethz.ch}
\affiliation{%
  \institution{ETH Zurich}
  \city{Zurich}
  \country{Switzerland}
}

\author{Sebastian Burmester}
\email{sburmester@student.ethz.ch}
\affiliation{%
  \institution{ETH Zurich}
  \city{Zurich}
  \country{Switzerland}
}

\author{Noa Sendlhofer}
\email{nsendlhofer@student.ethz.ch}
\affiliation{%
  \institution{ETH Zurich}
  \city{Zurich}
  \country{Switzerland}
}

\author{Jan Kamm}
\email{jakamm@student.ethz.ch}
\affiliation{%
  \institution{ETH Zurich}
  \city{Zurich}
  \country{Switzerland}
}

\author{Ruiheng Jiang}
\email{rjiang@student.ethz.ch}
\affiliation{%
  \institution{ETH Zurich}
  \city{Zurich}
  \country{Switzerland}
}

\author{Raffaello D'Andrea}
\email{rdandrea@ethz.ch}
\affiliation{%
  \institution{ETH Zurich}
  \city{Zurich}
  \country{Switzerland}
}

%% 3. Concise Header for Page Navigation
\renewcommand{\shortauthors}{Ramachandran et al.}

\begin{abstract}
Robotic choreography in open water is governed by nonlinear fluid dynamics, which impose significant challenges due to environmental disturbances and nonlinear system dynamics. This paper presents the cyber-physical architecture of Way of Water, a vertically integrated framework that orchestrates a fleet of autonomous surface vessels as a distributed choreographic platform. Moving beyond the surface-pixel paradigm, these vessels use laminar nozzles and multi-zone lighting to extend their expressive range from the 2D water plane into the 3D volumetric domain. Our primary contribution is the Way of Water Studio, a browser-based, timeline-compositing authoring paradigm that treats the fleet as a DAW-like instrument for music-responsive choreography. The Studio encapsulates Sequential Convex Programming for trajectory generation and Model Predictive Control for disturbance rejection presented through a visual timeline, broadening access to high-performance aquatic robotics for non-programmer artists. Grounding the Studio is the full cyber-physical stack: a custom holonomic chassis, a state-estimation and control stack tuned for the aquatic domain, and an LTE/MQTT fleet link with RTK-GPS time synchronization. We report on the system's validation across two distinct deployments: an 18-vessel Swan Lake interpretation at Lake Zurich and an 8-vessel Time Space Existence 2025 Venice Biennale demonstration at Forte Marghera, establishing a foundational reference for the design and deployment of fluidic robotic swarms.
\end{abstract}




\begin{teaserfigure}
  \includegraphics[width=\textwidth]{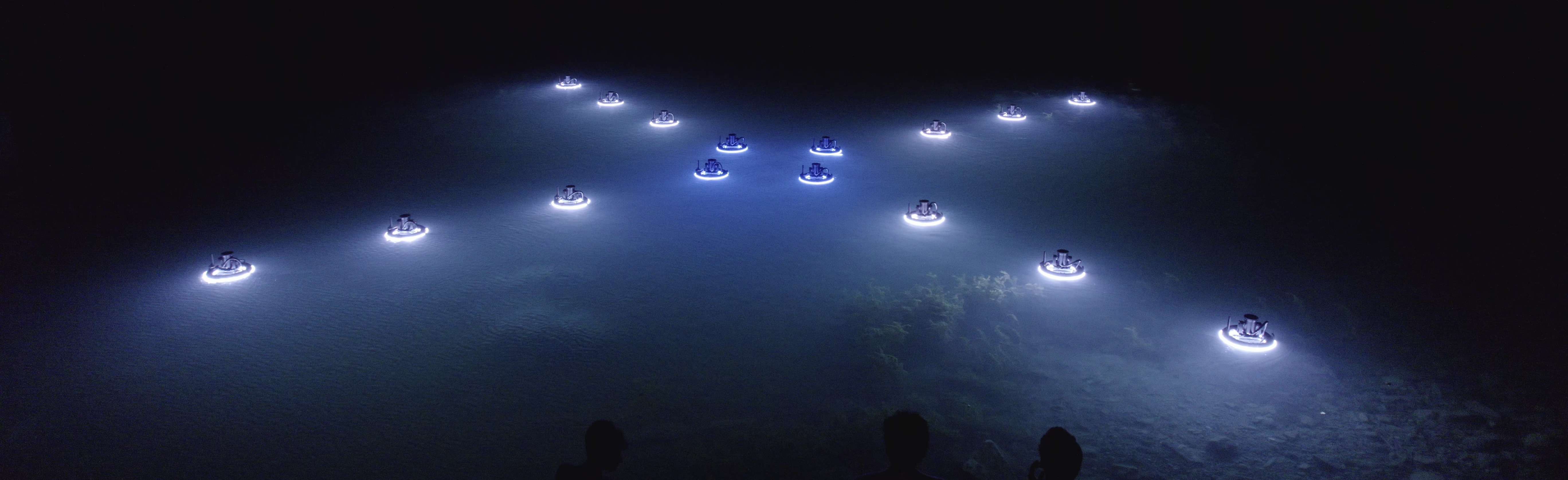}
  \caption{The Way of Water installation: A fleet of illuminated vessels performing synchronized motion on Lake Zurich.}
  \Description{A long-exposure photograph showing 24 illuminated robotic vessels on a dark body of water. The vessels leave light trails that form organic, fluid shapes, demonstrating synchronized movement. The background features the historic buildings of the Venice Arsenal at night.}
  \label{fig:teaser}
\end{teaserfigure}

\maketitle

\section{Introduction: Robotic Agency in Fluid Environments}
The domain of robotic art sits at the intersection of electromechanical precision and aesthetic expression. Specifically, the challenge of music-responsive choreography, where physical agents articulate the structural and emotional content of sound, requires a level of synchronization that is difficult to achieve in stochastic environments. While aerial swarms have successfully colonized the sky as "pixels in the sky" \cite{waibel2017drone}, surface-based aquatic systems struggle significantly due to the rigorous engineering required to operate in a boundary layer. The water surface imposes nonlinear disturbances such as wave dynamics, currents, and drift, which actively resist precise choreography. We present a framework that resolves these constraints, transforming a fleet of autonomous surface vessels (ASVs) into a distributed choreographic platform of \emph{dynamical pixels}. The vessels are treated as compositional primitives, with their position, heading, light state, and water-jet output all addressable from a single timeline track.

This paper establishes the cyber-physical architecture of Way of Water and, centrally, the authoring paradigm through which artists compose for it. To achieve the fidelity required for musically coherent motion, we analyze the mechanical engineering required to sustain precise station keeping, velocity tracking, and synchronized motion. We detail the adaptation of control strategies, Model Predictive Control (MPC) and Sequential Convex Programming (SCP), which are essential not only for navigation but also for the autonomous, collision-free choreography required to align motion with musical structure.

\subsection{The Narrative Arc of Aquatic Robotics}
The history of performance robotics has evolved from individual automatons to coordinated swarms. Aquatic environments offer a complementary medium to aerial shows, characterized by high endurance, intrinsic safety, and the evolution of collective behaviors in real-world settings \cite{duarte2016evolution}.

The Way of Water addresses the unique challenges of this medium through a vertically integrated approach. The vessels are not repurposed survey drones but purpose-built aesthetic platforms. Every design decision, from the holonomic X-drive propulsion to the custom refractive optics of the nozzle, serves the dual purpose of robustness and expression, using the fluid medium as a visual element to give physical form to music.

\subsection{From Script to Studio}
The primary contribution of this work is the \emph{Way of Water Studio}: a browser-based, timeline-compositing authoring paradigm for aquatic robotic swarms. It describes the evolution of the authoring workflow from a code-centric Python scripting framework toward a DAW-like interface that encapsulates the sophisticated mathematics of swarm coordination physics constraints, optimization solvers, and audio-structural mapping behind a visual timeline familiar to artists and musicians. 
This progression addresses the primary bottleneck in robotic performance: the friction between engineering rigidity and artistic fluidity. The Studio serves as an interface between the control parameters and the choreographer's musical intent, abstracting path-planning algorithms, collision constraints, and beat mapping behind a visual timeline. By documenting the complete system, from the hardware chassis to the web-based planning architecture, this work provides the necessary grounding for future research into distributed software architectures and real-time, musically responsive swarm coordination.

\begin{figure*}[t]
    \centering
    \includegraphics[width=\textwidth]{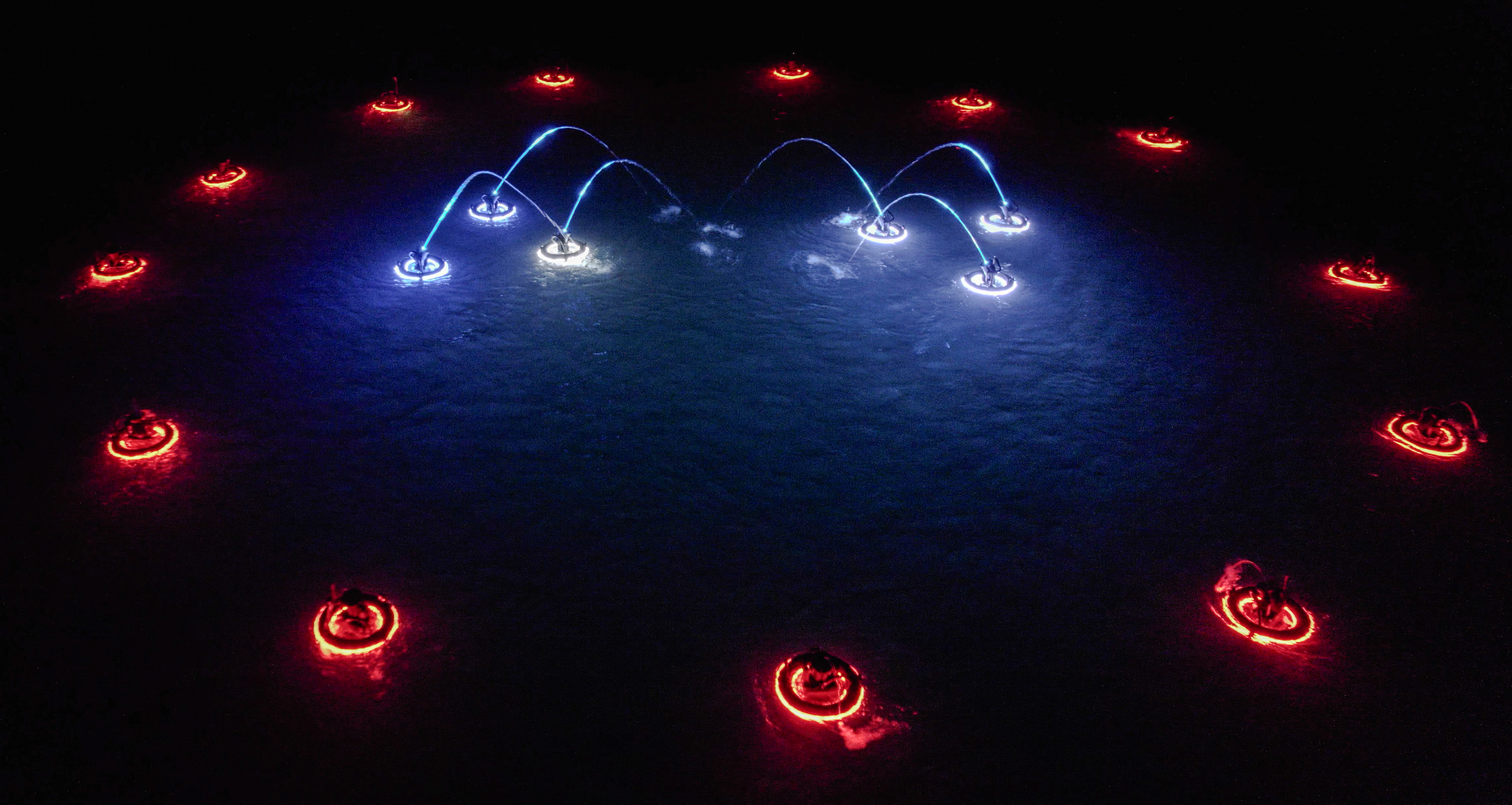}
    \caption{The Swan Lake performance, where autonomous vessels enact a balletic choreography, mirroring the elegance of the original score.}
    \Description{A high-angle shot of the robotic fleet in a circular formation on the water. The robots are emitting blue light, creating a stark contrast against the dark water. The formation appears precise and ordered.}
    \label{fig:swanlake}
\end{figure*}

\section{Related Work: The Genealogy of Performative Engineering}
The intersection of robotics and the arts has evolved from a niche curiosity into a rigorous discipline of performative engineering. We categorize the Way of Water within this lineage, tracing its roots from early cybernetics to modern distributed swarms.

\subsection{Foundational Theory: From Feedback to Swarms}
The genealogy of robotic art is rooted in the feedback loops of Cybernetic Serendipity (1968) and Gordon Pask's Conversation Theory, which posits that intelligence emerges from interaction \cite{pask1969architectural}. The shift to collective behaviors was formalized by Reynolds' Boids \cite{reynolds1987flocks}, and carried into the artistic domain by practitioners such as Bill Vorn, whose \emph{Hysterical Machines} and \emph{Grace State Machines} treat collective robotic agitation as performative dramaturgy \cite{vorn2017hysterical}, and Ken Rinaldo, whose \emph{Autopoiesis} (2000) demonstrated a gallery-scale ecology of robotic sculptures governed by local rules \cite{rinaldo2000autopoiesis}. An aquatic lineage runs in parallel: Gilberto Esparza's \emph{Plantas N\'omadas} \cite{esparza2010plantasnomadas} couples autonomous robotic-plant hybrids to polluted waterways; John McCormack's \emph{Holon} \cite{mccormack2019holon} stages an artificial aquatic ecosystem of reactive agents; and Audrey, Kelly, and St-Aubin's \emph{Vessels} \cite{audrey2019vessels} deploys autonomous craft as performative sculpture. Waibel et al.\ popularized swarm behaviors as pixels in the sky for aerial shows \cite{waibel2017drone}, and Duarte et al.\ demonstrated real-world aquatic swarms under evolutionary constraints \cite{duarte2016evolution}. Way of Water extends this aquatic-art genealogy into high-performance, timeline-composed swarm choreography, while deliberately deferring Pask/Reynolds-style local-rule emergence as a future primitive class within the Studio.

\subsection{Control Theoretic Frameworks}
To translate musical intent into physical motion, the system must enforce strict temporal and spatial compliance against a chaotic fluid medium. Control theory serves as the bridge between the choreographer's idealized score and the physical reality of hydrodynamics. Without rigorous state estimation and trajectory optimization, the swarm's musicality and its ability to synchronize with a transient or maintain a formation would be lost to nonlinear dynamics and environmental disturbances.

Precise aquatic coordination begins with robust \textbf{state estimation}; we employ an Extended Kalman Filter (EKF) fused with vessel-specific parameter identification to capture the system's hydrodynamic reality. Building on this foundation, high-performance choreography requires advanced strategies to handle the high-dimensional state space of a swarm.

\textbf{Collision Avoidance:} Augugliaro et al. pioneered SCP for aerial choreography \cite{augugliaro2012generation}, utilizing it to resolve collision-free trajectories for agile quadrotors in 3D space. We adapt this formulation to the planar aquatic domain, reducing the problem dimensionality and removing the jerk constraints on acceleration necessitated by aerial dynamics. \textbf{Fleet Transition:} To transition the fleet, the Linear Sum Assignment Problem (LSAP) minimizes total travel distance \cite{burkard2009assignment}, serving as a first layer of conflict resolution. \textbf{Tracking and Disturbance Rejection:} To proactively track, reject disturbance and drift, we employ MPC, optimizing control inputs over a finite horizon \cite{morgan2014mpc}. 

\subsection{Authoring Ecosystems and Robotic Expression}
The paradigm has shifted from choreography by code to choreography by design, evidenced by LLM-driven tools like SwarmGPT \cite{schuck2025swarmgpt} and physical interfaces like Zooids \cite{legoc2016zooids}. Recent work expands this scope through distributed sensor networks~\cite{wicaksono2024knitwork,wicaksono2022tapis} and AI-augmented instruments~\cite{blanchard2024symbiotic}.

This lineage traces back to Weinberg's Robotic Musicianship, where robots listen and improvise~\cite{weinberg2006toward,weinberg2023happy,weinberg2025dore}. Way of Water extends this agency from musical to spatial interaction: the vessels function as dynamical pixels~\cite{wang2020roboat} and interactive paintbrushes~\cite{santos2020interactive}, strictly governed by physics to visualize Newton's laws as aesthetic output. Recent MIR research using deep models such as Demucs~\cite{rouard2023hybrid} and CREPE~\cite{kim2018crepe} offers a forward path for driving procedural behavior from high-level musical features; the current Studio relies on the classical, offline-deterministic pipeline described in~\ref{sec:audio} and treats deeper MIR integration as future work.

Ultimately, Way of Water synthesizes robotics, physics-based modeling, and control theory into a unified framework for timeline-authored music-responsive choreography. 
While mathematical rigor ensures safety and coordination, it is the Way of Water Studio that serves as the critical abstraction layer, empowering artists to compose for aquatic swarms through a familiar visual timeline rather than code.

\section{The Physical Platform: Mechanical and Hydrodynamic Engineering}
The Way of Water vessel design balances several competing constraints: buoyancy, stability, agility, endurance, and visual impact. This section deconstructs the hardware specifications to provide a reference for the system's capabilities.

\begin{figure}[t]
    \centering
    \includegraphics[width=0.5\columnwidth]{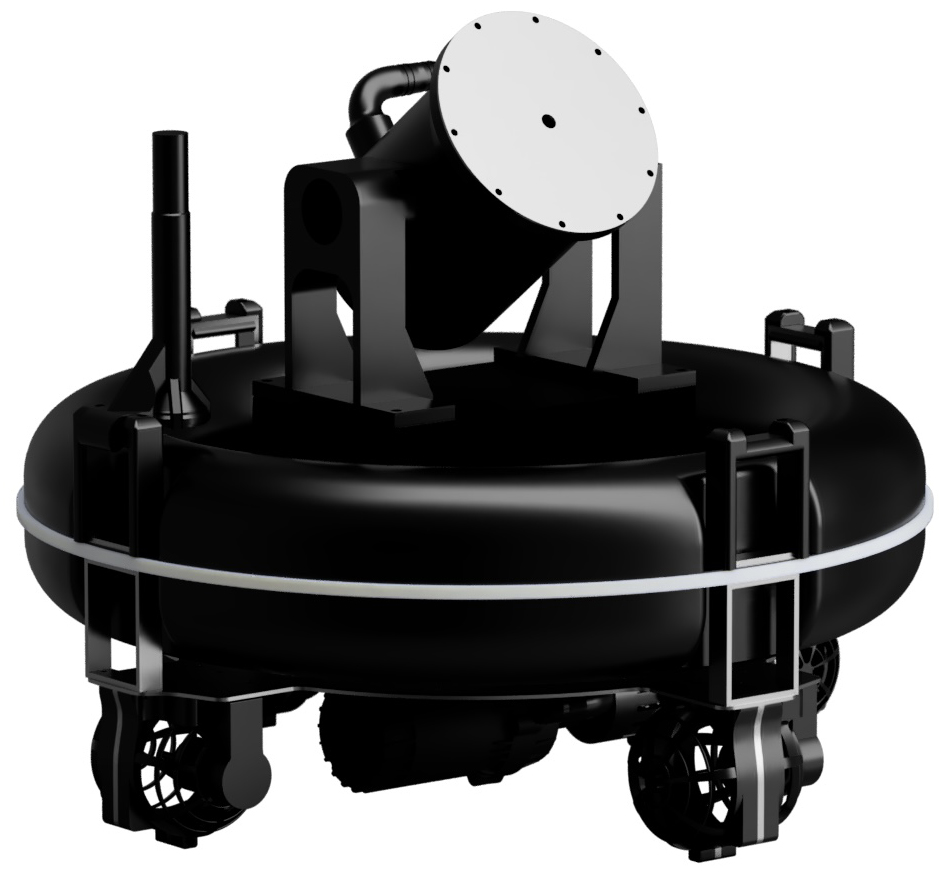}
    \caption{Overview of the hardware design, showing the integration of the lifebuoy chassis, propulsion system, and multi-layered lighting.}
    \Description{A diagram highlighting the components of the Way of Water vessel. Key parts labeled include the Lifebuoy Chassis, the 4-motor customized propulsion system, and the central E-Box. The vessel has an X-shaped frame atop a lifebuoy.}
    \label{fig:hardware_overview}
\end{figure}

\subsection{Holonomic Propulsion Architecture}
Unlike traditional rudder-steered boats, the Way of Water fleet utilizes a holonomic X-configuration with four brushless DC motors. This arrangement allows the vessel to translate in any direction ($x, y$) while simultaneously rotating ($r$) or holding a fixed heading to align the laminar fountain with artistic intent. Each thruster delivers 20N of thrust, allowing an operational speed of up to 0.8m/s. For safety, the thrusters are fully shrouded to enable operation in public waters.

\subsection{Optical Payloads: Laminar and Peripheral}
The vessel's visual signature relies on a custom laminar nozzle that generates a turbulence-free, internally illuminated water column acting as a volumetric pixel up to 6 meters in height. Complementing this vertical element, the chassis integrates a three-zone peripheral lighting architecture (Figure~\ref{fig:hardware_overview}) using individually addressable RGB LED strips an \textit{Internal Ring} that highlights the chassis structure, a \textit{Waterline Ring} that disperses streaks across the surface, and a \textit{Submersible Ring} that provides a diffuse underwater glow allowing independent control over the vessel's presence and its interaction with the fluid medium.

\subsection{Electronics and Power Architecture}
Operational endurance is a critical metric for public art. The fleet is powered by a custom Li-ion battery pack, enabling over 10 hours of continuous operation. The 25.2V architecture enables powerful thrusters, while a BMS provides active balancing and telemetry. The central electronics enclosure (E-Box), machined from aluminum, is sealed to IP67 and acts as a unified heatsink for the ESCs and battery. It houses a Raspberry Pi 5, providing significant onboard compute power for the decentralized execution of sophisticated solvers, such as OSQP for MPC, directly on the agent. The propulsion thrusters are fully submersible (IP68), and the external connectors are IP67-rated push-pull bayonet types; the platform has been operated for tens of hours per show across both freshwater (Lake Zurich) and brackish (Venice lagoon) deployments without sealing failures.

\subsection{Fleet Communication and Time Synchronization}
\label{sec:comms}
Each vessel carries an onboard LTE modem that streams telemetry and receives RTK-GPS corrections through an MQTT broker; an additional WiFi channel is used at the shore for high-bandwidth, low-cost communication during setup and mission upload. Fleet-wide show-clock synchronization is derived from the RTK-GPS pulse-per-second, providing sub-millisecond temporal alignment between vessels without requiring a dedicated master clock. The MQTT topic structure separates concerns for low-rate operator commands (start/abort/return-to-shore). Typical end-to-end planner-to-vessel latency over LTE is in the 80--250\,ms range; show choreography is therefore not closed-loop over the radio link. Show data is pre-uploaded over WiFi at the shore, then executed under the shared GPS timebase: a vessel can tolerate a complete LTE blackout for several minutes while still hitting its choreography to within sub-RTK precision, and the planner detects stale-clock or off-pattern vessels by monitoring the deviation between expected and reported state in \texttt{telemetry}.

\section{The Control Theoretic Framework: Taming the Stochastic Environment}
\label{sec:control}
The operational stability of the Way of Water relies on the suppression of environmental disturbances. To achieve smooth trajectories in the presence of wind and current, the control stack utilizes established state estimation and predictive control techniques adapted for the aquatic domain.

% \newpage

\subsection{Sensor Fusion and Heading Estimation}
To enable precise choreography, the system must robustly estimate each vessel's full 3-DOF state. We employ a Multi-Rate EKF with a state vector encompassing position, velocity, and orientation: $\mathbf{x} = [x, y, \psi, u, v, r]^T$.~\cite{julier1997new} 
\textbf{RTK-GPS:} Provides position ($x, y$) and velocity ($v_x, v_y$) at 10Hz with centimeter-level precision.
\textbf{IMU:} Provides angular rates ($\omega$) and linear accelerations ($a$) at 100Hz, filling the gaps between GPS updates.
Crucially, the system determines heading ($\psi$) through an observability-based estimator. By correlating the known control inputs with the observed inertial displacement from RTK-GPS, the EKF converges on the true orientation. This sensor fusion strategy ensures accurate, drift-free tracking of the vessel's pose in an open-water environment.

\subsection{Model Predictive Control}
While reactive PID controllers are sufficient for station-keeping, they lack the anticipatory behavior required for fluid, organic motion in a high-inertia medium. Pure feedforward control, conversely, is prone to drift from wind and current.

We employ MPC to solve a finite-horizon optimization problem at 10Hz. This allows the vessel to anticipate turns, lean into maneuvers to counter inertia, and exhibit inherent recovery behaviors if a wave knocks it off course.

\textbf{Objective Function:} The cost function $J$ minimizes deviation from the choreographed trajectory. The state-tracking weight $\mathbf{Q}$ is tuned to prioritize heading precision (keeping laminar nozzles strictly parallel for visual coherence) while allowing mild positional compliance to absorb wave energy; the input-tracking weight $\mathbf{R}$ penalizes deviation from the feedforward command to suppress jitter.

\begin{equation}
    J = \sum_{k=0}^{N-1} [(\mathbf{x}_k - \mathbf{x}_{r,k})^T \mathbf{Q} (\mathbf{x}_k - \mathbf{x}_{r,k}) + (\mathbf{u}_k - \mathbf{u}_{r,k})^T \mathbf{R} (\mathbf{u}_k - \mathbf{u}_{r,k})]
\end{equation}

\begin{figure}[htbp]
    \centering
    \includegraphics[width=0.47\textwidth]{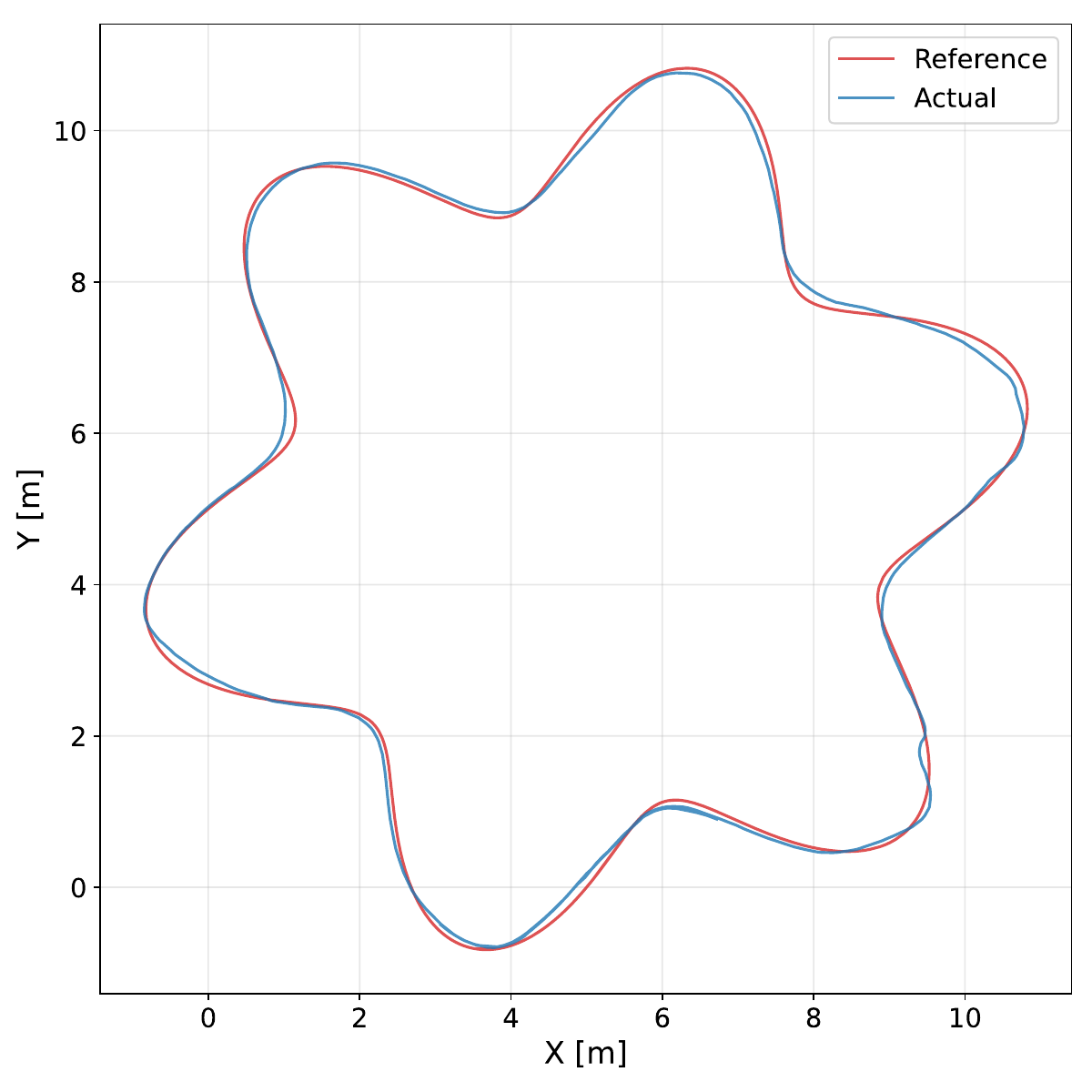}
    \caption{MPC tracking performance under wave and current disturbance. The controller maintains high-precision station keeping, achieving a RMSE of \textbf{0.038\,m} in position ($x, y$) and \textbf{1.4$^{\circ}$} in heading ($\psi$).}
    \Description{Plot showing the MPC tracking performance with low RMSE in position and heading under disturbances.}
    \label{fig:trajectory_comparison}
\end{figure}

Table~\ref{tab:mpc_symbols} summarises the symbols. OSQP solves the resulting QP in real time on the Raspberry Pi. In the field, this controller delivers per-vessel RMSE of \textbf{0.038\,m} in cross-track position $(x,y)$ and \textbf{1.4$^{\circ}$} in heading $\psi$ (Figure~\ref{fig:trajectory_comparison}). These figures are stable across fleets of 8 (Venice Biennale) and 18 (Lake Zurich) vessels under sustained wind and wave conditions, the operational envelope tested during the deployments. 
%Aggregating onboard EKF telemetry from the Lake Zurich Swan Lake show (18 vessels, $\sim$60\,m show area, $\sim$ 20\,min of recorded run), per-vessel cross-track error was \textcolor{red}{[mean$\pm$std]}\,m and heading error \textcolor{red}{[mean$\pm$std]}$^{\circ}$, with the largest excursions clustered around fleet transitions rather than station-keeping segments; wind during recording ranged \textcolor{red}{[range]}\,m/s. 
Equivalent post-hoc telemetry was not retained for Forte Marghera, so we report cross-deployment robustness only at the qualitative level. Beyond the tested envelope, performance degrades primarily through GPS-correction integrity rather than control bandwidth.

\begin{table}[ht]
\centering
\footnotesize
\caption{MPC formulation symbols.}
\label{tab:mpc_symbols}
\begin{tabularx}{\columnwidth}{l X}
\toprule
Symbol & Meaning \\
\midrule
$N$ & prediction horizon (steps) \\
$\mathbf{x}_k,\,\mathbf{u}_k$ & predicted state and control input at step $k$ \\
$\mathbf{x}_{r,k},\,\mathbf{u}_{r,k}$ & reference trajectory and feedforward \\
$\mathbf{Q},\,\mathbf{R}$ & state and input weighting (heading-prioritised) \\
\bottomrule
\end{tabularx}
\end{table}

\subsection{Collision Avoidance via SCP}
Generating safe trajectories for a fleet of 24 robots is a non-convex optimization problem. We utilize SCP to solve this offline.

The pairwise collision avoidance constraint between robot $i$ and robot $j$ is defined as:
\begin{equation}
    || \mathbf{p}_i[k] - \mathbf{p}_j[k] ||_2 \ge R_{safe}
\end{equation}
where $\mathbf{p}[k]$ is the position at time step $k$ and $R_{safe}$ is the minimum safety radius (hull + buffer). Since this constraint is non-convex, SCP linearizes it around a candidate trajectory $\bar{\mathbf{p}}$ from the previous iteration. The resulting convex constraint is:
\begin{equation}
    || \bar{\mathbf{d}}_{ij}[k] ||_2 + \frac{\bar{\mathbf{d}}_{ij}[k]^T}{|| \bar{\mathbf{d}}_{ij}[k] ||_2} (\mathbf{d}_{ij}[k] - \bar{\mathbf{d}}_{ij}[k]) \ge R_{safe}
\end{equation}
where $\mathbf{d}_{ij} = \mathbf{p}_i - \mathbf{p}_j$ is the relative position vector. This formulation defines a separating hyperplane between agents. The solver iterates this linear approximation until the trajectory converges to a solution that satisfies the original non-convex constraints, guaranteeing mathematically collision-free paths prior to deployment.

\section{The Way of Water Studio}
The Way of Water Studio is the primary contribution of this work: a browser-based, DAW-like authoring environment that treats a fleet of ASVs as a single timeline-composed instrument for music-responsive choreography (Figure~\ref{fig:motion_editor}). It transitions the workflow from code-based scripting to visual, timeline-based compositing, abstracting the rigorous control theory of Section~\ref{sec:control} behind an interface designed for rapid artistic iteration. Instead of manipulating individual waypoints, choreographers define trajectories parametrically; these are pre-generated and validated against the fleet's dynamic constraints to ensure physical feasibility before upload.

The Studio is a \emph{timeline-authored choreography platform with audio-reactive payload modulation}, validated through two real-world case studies. It is not a real-time improvisatory instrument, nor an autonomous music-listening agent, nor a system in which motion itself is generated directly from audio. Vessel motion is composed by the choreographer against a pre-analyzed audio reference; the audio pipeline (Section~\ref{sec:audio}) drives lights and water-jet payloads. Closing the gap between music-driven \emph{payload} and music-driven \emph{motion}, together with a more formal evaluation program, is the natural next research direction; the contribution here is the authoring paradigm and the cyber-physical stack that make such future work tractable.

\subsection{Design Requirements: The Swan Lake Prototype}
\label{sec:swan_lake}
\begin{figure}[t]
    \centering
    \includegraphics[width=0.7\columnwidth]{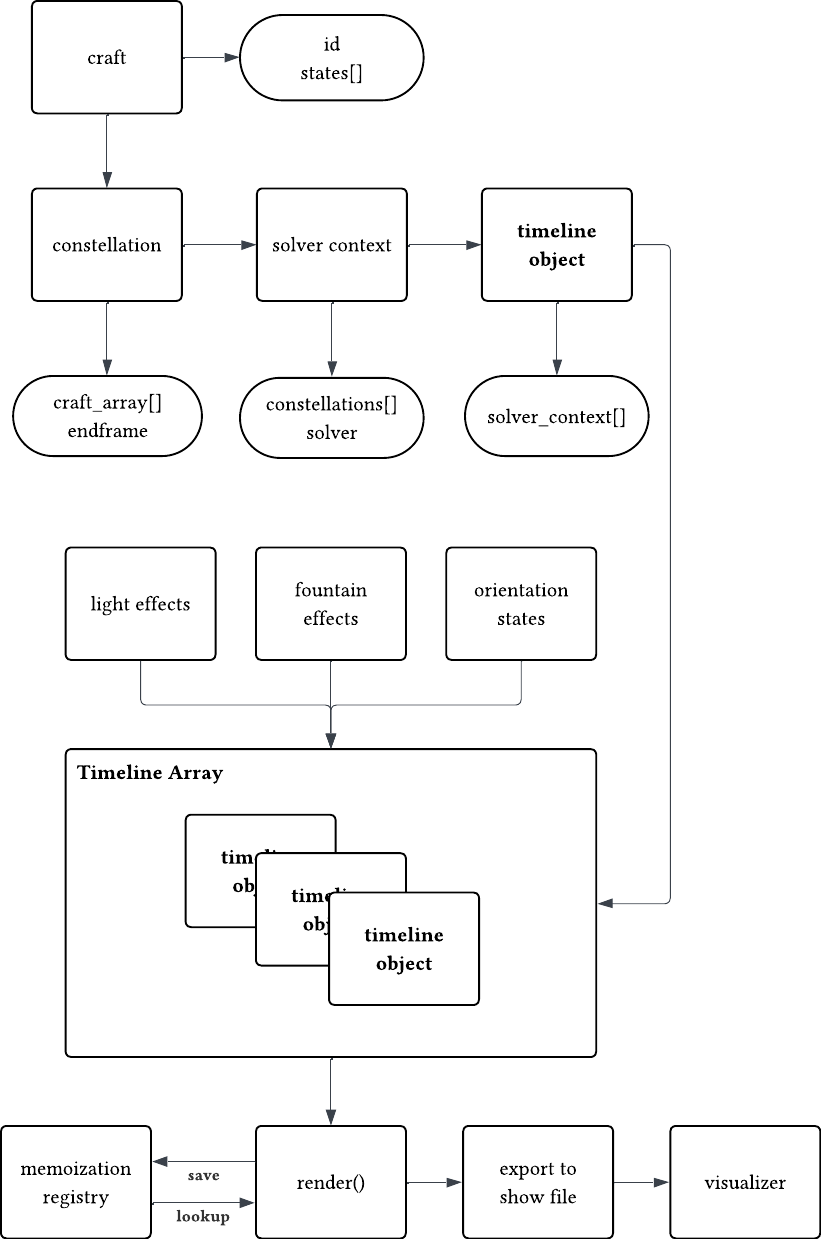}
    \caption{Architecture of the code-based planner used for Swan Lake. The system decouples the artistic Script from the Solver Context, validating physical feasibility against the dynamics model.}
    \Description{Architecture of the code-based planner used for Swan Lake. The system decouples the artistic Script from the Solver Context, validating physical feasibility against the dynamics model.}
    \label{fig:planner_arch}
\end{figure}

Prior to the development of the graphical Studio, the choreography for the Swan Lake performance was authored using a code-first approach. This framework, illustrated in Figure \ref{fig:planner_arch}, served as the foundational prototype for the current system. It relied on a scripted pipeline where choreographers defined trajectories using Python primitives. While rigorous and flexible, this approach lacked real-time visual feedback; it required specific programming knowledge and a slow compilation-simulation loop to validate even minor adjustments.

Insights from the Swan Lake deployment defined the architectural requirements for the new Studio, resulting in a system composed of three primary interaction layers: a timeline-based compositing interface, parametric primitives for behavior definition, and a real-time 3D viewport for rapid artistic iteration.

\subsection{The Timeline Interface: Timeline-Based Compositing}
\begin{figure*}[t]
    \centering
    \includegraphics[width=\textwidth]{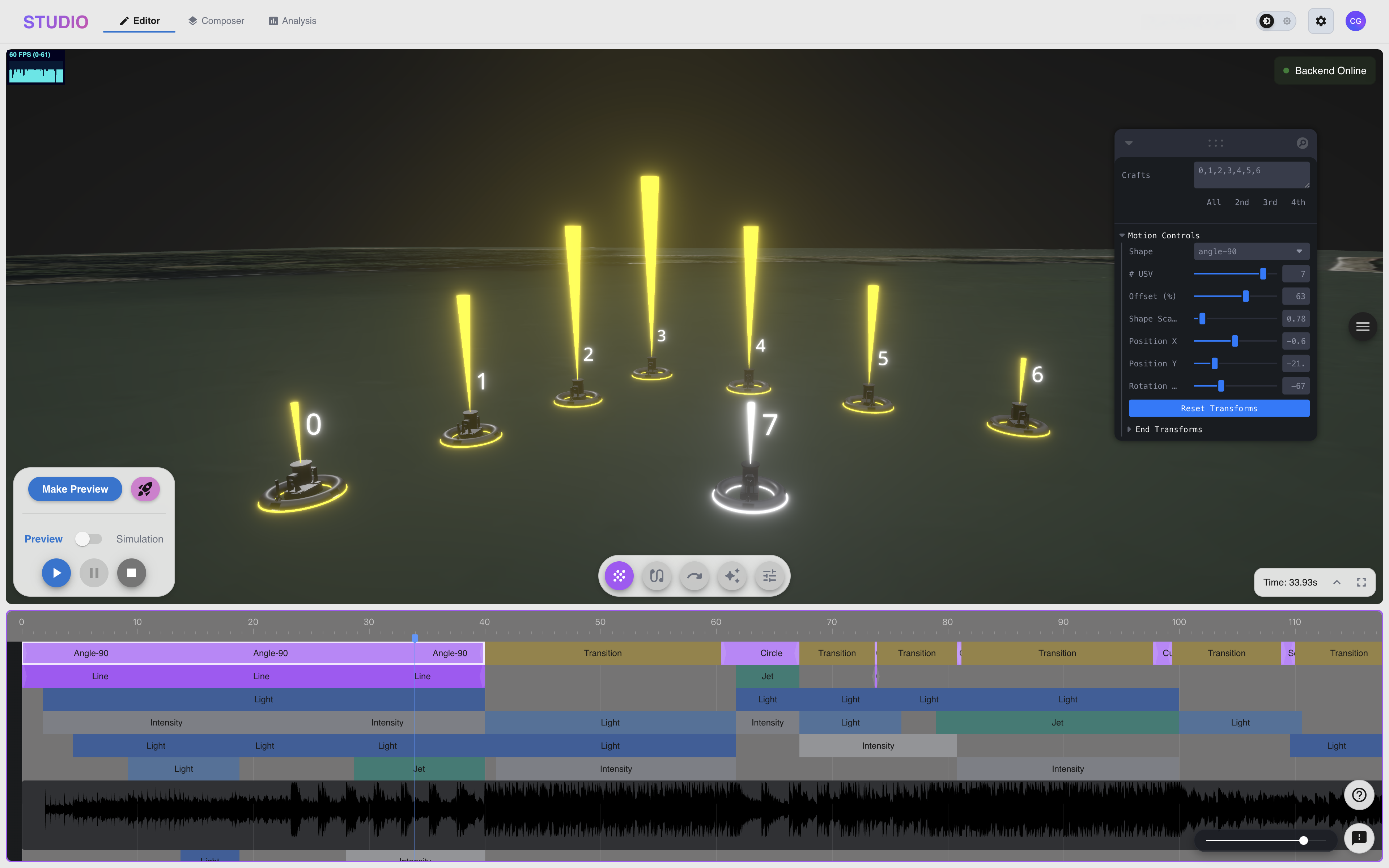}
    \caption{Parametric Motion Editor. Double-clicking the timeline (bottom) creates an effect, which can be dragged and resized. There are five effect types are Motion, Transition, Jet, Light and Intensity, which control movements and payloads. Upon selecting an effect, its parameters can be tweaked (floating panel). The 3D viewport (center) previews the choreography.}
    \Description{A screenshot of the Motion Editor panel in WOW Studio. It shows parameters for defining a formation transition, such as Start Shape and End Shape, along with duration adjustments. A visual preview of the trajectory logic is implied.}
    \label{fig:motion_editor}
\end{figure*}

Mirroring the layout of Digital Audio Workstations (DAWs), the Studio adopts a timeline-based compositing paradigm. This approach moves away from rigid waypoint lists, treating fleet behaviors as high-level parametric clips that can be dragged, resized, and layered non-destructively. As illustrated in Figure~\ref{fig:motion_editor}, the interface organizes these behaviors into synchronized sequencing layers:

Motion Layer: Defines the geometric topology of the fleet. As seen in the Motion Controls panel (Figure~\ref{fig:motion_editor}, right), users manipulate meta-parameters such as shape topology (e.g., circle, line), scale, and Cartesian offsets rather than individual robot coordinates. Transition effects transform the shapes from one to another, which can be created manually or using the LSAP-based solver.

Payload Layer: A track-agnostic sequencing layer that controls the vessel's peripherals. This includes the Light Track for LED arrays (color, intensity, effects like Rainbow Chase) and the Jet Track for actuating the laminar nozzles.

Audio Reference: A visual waveform serves as the temporal anchor (Figure~\ref{fig:motion_editor}, bottom), allowing choreographers to align fleet transitions and payload effects with musical transients with visual precision.

\subsection{Parametric Primitives}
\label{sec:lsap}
To bridge the gap between artistic intent and control constraints, the system relies on parametric primitives. Instead of manually positioning individual agents on a geometric shape, users define the meta-state of a formation.

For example, a circle primitive is not defined by 24 coordinate sets but by its global position ($x, y$), rotation ($\theta$), scale, and phase offset. The global position corresponds to the center of each primitive's bounding box. The phase offset shifts all crafts along the parameterized vector graphic path that defines the shape. 
By choosing this approach, where primitives are vector graphics that are normalized based on their bounding boxes, it allows for quick changes to shapes at a later stage, while the crafts retain their approximate location within the choreography.

\subsection{Transitions}
\label{sec:transitions}
Transitions are performed between the primitive shapes. 
The choreographer can manually match start and end positions or use the LSAP solver. 

The LSAP solver uses squared Euclidean distance as weights. For this approach, it has been shown that the minimum distance between trajectories is guaranteed to be $d_{min} = \sqrt{2}\delta/2 $ where $\delta$ is the minimum distance between any two crafts across the start and end positions \cite{kratky2025catoracollisionawaretimeoptimalformation, turpin2014capt}. For our crafts with a diameter of 50cm, this implies that a gap of $2*50 \text{cm} /\sqrt{2} - 50\text{cm} \approx 21\text{cm} $ is enough to guarantee collision-free paths.   
Note that this approach assumes that the crafts are identical and that each craft can assume any position in the target state. 
Manual adjustments of the LSAP results allow for artistic expression and desired sub-optimality. 

Adjusting the transition smoothness parameters directly alters the resulting velocity profiles, enabling the choreographer to align movement dynamics with expressive qualities such as calmness or intensity. Note that the SCP solver from the Swan Lake planner could be integrated to allow for more intricate transitions that are more complex than straight lines, and allow for fixed end position assignments of crafts.

\subsection{Audio Pipeline}
\label{sec:audio}
To bridge the gap between musical structure and robotic action, the framework integrates an \emph{offline} audio analysis pipeline powered by Librosa~\cite{mcfee2015librosa} and Scipy~\cite{virtanen2020scipy}. In the current Studio, choreographies are authored against a pre-analyzed audio track and rendered into a show file before deployment; live audio-driven operation is explicitly out of scope for this paper. We are also explicit about the division of labor between manual and audio-driven authoring: motion is composed in the timeline by the choreographer (with the audio waveform as the temporal reference), while the audio features described below primarily drive payload modulation of light intensity, jet height, and event triggers. Closing this gap so that motion itself becomes audio-reactive is a deliberate next step, not a hidden capability of the current system. The frontend Studio interface focuses on timeline visualization, while the backend generates the complete choreography state file by combining the processed audio signal with the effect data from the timeline.

Feature Extraction: Upon upload, each audio file is loaded at its native sample rate to preserve temporal fidelity. A Short-Time Fourier Transform (STFT) with a Hann window of 2048 samples and a hop size of 512 samples serves as the basis for all spectral analysis. The pipeline extracts the following feature sets:
{Rhythm features:} Global tempo estimation and beat tracking, onset detection using the onset strength envelope with subsequent peak picking, and onset strengths normalized to $[0, 1]$. \textbf{Energy features:} Root-mean-square (RMS) energy computed per frame, with global energy peaks. \textbf{Spectral features:} Spectral centroid (timbral brightness), spectral rolloff at the 85th percentile, and zero-crossing rate for texture characterization.

Frequency Band Decomposition: For audio-reactive effects, the system computes band-limited energy envelopes on demand. The STFT magnitude spectrogram is partitioned into seven perceptually motivated frequency bands Sub Bass (20--60\,Hz), Bass (60--250\,Hz), Low Mids (250--500\,Hz), Midrange (500--2000\,Hz), High Mids (2--4\,kHz), Presence (4--6\,kHz), and Air (6--20\,kHz) by summing magnitude bins within each range and normalizing to unit peak. Additionally, a custom frequency band can be specified. Per-band peak detection with configurable height, distance, and prominence thresholds allows designers to trigger choreographic events from specific spectral regions. These per-band results are cached independently, so repeated queries incur no recomputation.

Intensity Modulation: The extracted audio features drive a multi-stage intensity modulation chain that transforms base color values or water jet intensities into musically responsive output. 

Jumping and Flash Patterns: For spatially distributed effects, detected beats or frequency-band peaks serve as trigger events. A \emph{jumping} pattern sequentially or randomly activates individual crafts at each trigger, with per-craft intensity decaying between activations. The inverse \emph{flash} pattern maintains all crafts at peak intensity and introduces an anticipation dim before each trigger, creating a rhythmic breathing effect synchronized to the music's temporal structure.

% This modular framework allows for straightforward adaptation to new payloads or further desired effects by exposing the results from the audio analysis to the individual effect processing pipelines. 

\subsection{Authoring Workflow and Deployments}
\label{sec:venice_biennale}
The workflow is designed to minimize cognitive load while maximizing control authority. Users begin by initializing a project and uploading the master audio track; the physical fleet configuration (number of active USVs, show area bounds) is then set so the planner can enforce the specific hardware and geometric constraints. By abstracting the underlying robotic control layers, the system lets artists focus on the expression of music through light, motion, and fluid dynamics.

Comparative authoring assessment: We conducted an informal comparative assessment across two production cycles. The Swan Lake performance (18 vessels across an $\sim$60\,m show area on Lake Zurich) was authored entirely through the code-first pipeline, which restricted authorship to two engineers with Python fluency and required 4-8 hours per scene with a 5-minute script--compile--simulate iteration loop. The Lake Zurich demonstration with 18 crafts and the Venice Biennale demonstration (8 vessels in the smaller lagoon footprint at Forte Marghera) were produced entirely within the Studio; authoring time dropped to 90-120 minutes per song, and the edit-to-preview cycle to under 10 seconds. Most significantly, the Studio eliminated the requirement for programming expertise, expanding the authoring team. These observations stem from a single production team and are preliminary; a formal usability evaluation with external choreographers and musicians remains important future work.

Failure modes and aesthetic resilience: Across both deployments, no catastrophic failure has occurred to date, but two incidents are worth documenting. In one performance, a single vessel lost its show-clock synchronization and continued executing its full program out of phase; rather than reading as a fault, this manifested as a formation asymmetry that audience members interpreted as an intentional perturbation, an unintended but illustrative form of choreographic noise. In another case, a single vessel suffered a mechanical fault mid-show; its lights were remotely disabled, and it drifted out of the active formation unnoticed. 

Informal audience and collaborator reflections: Collaborator responses during the Venice Biennale run most often framed the work in terms of the medium itself \emph{``water dancing to music, with lights and motion.''} A recurring informal observation was a collective hush in the spectator field: children who had been playing loudly prior to a show would fall silent for the duration of the choreography and resume only afterward. We report these reactions as anecdotal rather than evaluative; systematic audience and artist studies, particularly with choreographers and composers external to the development team, are a future step.

\section{Conclusion}
We have presented the Way of Water: a vertically integrated cyber-physical framework for music-responsive choreography on open water with an authoring paradigm, the Way of Water Studio, that treats a fleet of ASVs as a timeline-composed instrument rather than as a programming problem. The Studio encapsulates SCP for collision avoidance, the LSAP for fleet transitions, and MPC for disturbance rejection behind a DAW-like timeline, allowing choreographers with musical rather than engineering backgrounds to compose for a distributed aquatic swarm.

Two deployments demonstrate versatility: the 18-vessel Swan Lake interpretation on Lake Zurich confirmed precision in station-keeping and laminar stability, while the 8-vessel Venice Biennale demonstration at Forte Marghera proved robustness in open water under agile, high-velocity maneuvers. The project sits at the intersection of several communities: engineers, choreographers, musicians, and public audiences, whose convergence is itself part of the contribution.
%: lowering the authoring threshold expanded the in-house team from two engineer-programmers to a five-person interdisciplinary group with musical backgrounds, and 
The Biennale spectator field foregrounded a public reading of robotic choreography as an ecological-aesthetic spectacle rather than a technical demonstration. We see the Studio as a step toward making such interdisciplinary, audience-facing work reproducible by other teams.

Several directions remain open. \emph{Scalability:} the current centralized solver practically limits the fleet to $\sim$24 vessels; larger swarms will require distributed middleware and hierarchical planning. \emph{Site and environment:} the installation relies on fixed audio infrastructure and is sensitive to weather and seasonal visibility; distributed audio streaming and more robust perception are needed to expand the operational envelope. \emph{Choreographic primitives:} current primitives are top-down parametric formations; a natural extension is a class of Reynolds/Pask-style local-rule primitives that would let the Studio host bottom-up emergent behaviors alongside scripted ones. 
%Architecturally, such a primitive would expose the same start/end/parameter interface as a shape primitive, but its parameters would specify a per-vessel rule (e.g., \emph{follow nearest neighbor at distance $d$}) evaluated locally at runtime, with the existing SCP layer retained as a collision-safety guard. 
This keeps the Studio's compositional grammar uniform while opening an emergent design space. \emph{Live operation:} the current audio pipeline is offline; extending it to real-time listening and improvisation would bring the platform closer to the lineage of robotic musicianship. \emph{Evaluation:} formal usability studies with external choreographers and audience-experience research remain the most important immediate next step.

\section*{Media Links}

\begin{itemize}
	\item Skyfall: \url{https://youtu.be/gprHw5OZqsQ}
	\item The Most Beautiful Boy: \url{https://youtu.be/VHYqJ1hxct0}
  \item Forte Marghera: \url{https://youtu.be/G4cM6xbG7PA}
\end{itemize}

\section*{Ethical Standards}

This work was supported by ETH Zurich. The authors declare no competing financial interests or conflicts of interest regarding the publication of this paper. Motivated by the need for sustainable, reusable alternatives to pyrotechnics, this research prioritizes environmental stewardship. To minimize ecological impact, the Way of Water fleet utilizes zero-emission electric propulsion and fully shrouded impellers, preventing chemical contamination and physical injury to aquatic wildlife. Operational safety in public venues (Lake Zurich, Venice Lagoon) was guaranteed through a multi-layered control architecture, where all autonomous trajectories were validated via Software-in-the-Loop (SITL) simulation prior to physical execution. The evaluation of the authoring workflow relied solely on the internal production team's professional practice; as no external human subjects were recruited, informed consent procedures were not applicable, and formal Institutional Review Board (IRB) approval was not required.

\begin{acks}
We extend our gratitude to our technical staff Daniel Wagner, and Matthias Müller, as well as the Focus Project 2025 participants Jonathan Nowack, Fjodor Lundgren, Janis Weyrich, and Daniel Gelfenbaum for their hard work and continued support during the development of the fleet.
\end{acks}

% References
\balance
\bibliographystyle{ACM-Reference-Format}
\bibliography{sample-references}


\begin{thebibliography}{29}

%%% the Web address.

\ifx \showCODEN    \undefined \def \showCODEN     #1{\unskip}     \fi
\ifx \showDOI      \undefined \def \showDOI       #1{#1}\fi
\ifx \showISBNx    \undefined \def \showISBNx     #1{\unskip}     \fi
\ifx \showISBNxiii \undefined \def \showISBNxiii  #1{\unskip}     \fi
\ifx \showISSN     \undefined \def \showISSN      #1{\unskip}     \fi
\ifx \showLCCN     \undefined \def \showLCCN      #1{\unskip}     \fi
\ifx \shownote     \undefined \def \shownote      #1{#1}          \fi
\ifx \showarticletitle \undefined \def \showarticletitle #1{#1}   \fi
\ifx \showURL      \undefined \def \showURL       {\relax}        \fi
% invisible to TeX
\providecommand\bibfield[2]{#2}
\providecommand\bibinfo[2]{#2}
\providecommand\natexlab[1]{#1}
\providecommand\showeprint[2][]{arXiv:#2}

\bibitem[Audry et~al\mbox{.}(2019)]%
        {audrey2019vessels}
\bibfield{author}{\bibinfo{person}{Sofian Audry}, \bibinfo{person}{Stephen
  Kelly}, {and} \bibinfo{person}{Samuel St-Aubin}.}
  \bibinfo{year}{2019}\natexlab{}.
\newblock \bibinfo{title}{Vessels}.
\newblock \bibinfo{howpublished}{Robotic art installation}.
\newblock
\urldef\tempurl%
\url{https://sofianaudry.com/works/vessels/}
\showURL{%
\tempurl}
\newblock
\shownote{A performative installation of autonomous aquatic vessels exhibited
  in artistic venues.}.


\bibitem[Augugliaro et~al\mbox{.}(2012)]%
        {augugliaro2012generation}
\bibfield{author}{\bibinfo{person}{Federico Augugliaro},
  \bibinfo{person}{Angela~P Schoellig}, {and} \bibinfo{person}{Raffaello
  D'Andrea}.} \bibinfo{year}{2012}\natexlab{}.
\newblock \showarticletitle{Generation of collision-free trajectories for a
  quadrocopter fleet: A sequential convex programming approach}. In
  \bibinfo{booktitle}{\emph{2012 IEEE/RSJ International Conference on
  Intelligent Robots and Systems (IROS)}}. IEEE, \bibinfo{publisher}{IEEE},
  \bibinfo{address}{New York, NY, USA}, \bibinfo{pages}{1917--1922}.
\newblock


\bibitem[Blanchard et~al\mbox{.}(2024)]%
        {blanchard2024symbiotic}
\bibfield{author}{\bibinfo{person}{Lancelot Blanchard}, \bibinfo{person}{Perry
  Naseck}, \bibinfo{person}{Eran Egozy}, {and} \bibinfo{person}{Joseph~A
  Paradiso}.} \bibinfo{year}{2024}\natexlab{}.
\newblock \bibinfo{booktitle}{\emph{Developing Symbiotic Virtuosity:
  AI-Augmented Musical Instruments and Their Use in Live Music Performances}}.
\newblock \bibinfo{type}{{T}echnical {R}eport}. \bibinfo{institution}{MIT Media
  Lab}, \bibinfo{address}{Cambridge, MA, USA}.
\newblock
\urldef\tempurl%
\url{https://doi.org/10.21428/e4baedd9.69c11de7}
\showDOI{\tempurl}
\newblock
\shownote{An MIT Exploration of Generative AI}.


\bibitem[Burkard et~al\mbox{.}(2009)]%
        {burkard2009assignment}
\bibfield{author}{\bibinfo{person}{Rainer Burkard}, \bibinfo{person}{Mauro
  Dell'Amico}, {and} \bibinfo{person}{Silvano Martello}.}
  \bibinfo{year}{2009}\natexlab{}.
\newblock \bibinfo{booktitle}{\emph{Assignment Problems}}.
\newblock \bibinfo{publisher}{SIAM}, \bibinfo{address}{Philadelphia, PA}.
\newblock


\bibitem[Duarte et~al\mbox{.}(2016)]%
        {duarte2016evolution}
\bibfield{author}{\bibinfo{person}{Miguel Duarte}, \bibinfo{person}{Vasco
  Costa}, \bibinfo{person}{Jorge Gomes}, \bibinfo{person}{Tiago Rodrigues},
  \bibinfo{person}{Fernando Silva}, \bibinfo{person}{Sancho~M Oliveira}, {and}
  \bibinfo{person}{Anders~Lyhne Christensen}.} \bibinfo{year}{2016}\natexlab{}.
\newblock \showarticletitle{Evolution of collective behaviors for a real swarm
  of aquatic surface robots}.
\newblock \bibinfo{journal}{\emph{PLoS One}} \bibinfo{volume}{11},
  \bibinfo{number}{3} (\bibinfo{year}{2016}), \bibinfo{pages}{e0151834}.
\newblock


\bibitem[Esparza(2010)]%
        {esparza2010plantasnomadas}
\bibfield{author}{\bibinfo{person}{Gilberto Esparza}.}
  \bibinfo{year}{2010}\natexlab{}.
\newblock \bibinfo{title}{Plantas N{\'o}madas}.
\newblock \bibinfo{howpublished}{Artwork installation}.
\newblock
\urldef\tempurl%
\url{https://gilbertoesparza.net/portfolio/plantas-nomadas/}
\showURL{%
\tempurl}
\newblock
\shownote{Autonomous hybrid robotic-plant organisms that migrate along polluted
  waterways. VIDA 13.2 Art and Artificial Life International Awards,
  Fundaci{\'o}n Telef{\'o}nica.}.


\bibitem[Julier and Uhlmann(1997)]%
        {julier1997new}
\bibfield{author}{\bibinfo{person}{Simon~J Julier} {and}
  \bibinfo{person}{Jeffrey~K Uhlmann}.} \bibinfo{year}{1997}\natexlab{}.
\newblock \showarticletitle{New extension of the Kalman filter to nonlinear
  systems}. In \bibinfo{booktitle}{\emph{Signal Processing, Sensor Fusion, and
  Target Recognition VI}}, Vol.~\bibinfo{volume}{3068}. SPIE,
  \bibinfo{publisher}{SPIE}, \bibinfo{address}{Bellingham, WA, USA},
  \bibinfo{pages}{182--193}.
\newblock


\bibitem[Kim et~al\mbox{.}(2018)]%
        {kim2018crepe}
\bibfield{author}{\bibinfo{person}{Jong~Wook Kim}, \bibinfo{person}{Justin
  Salamon}, \bibinfo{person}{Peter Li}, {and} \bibinfo{person}{Juan~Pablo
  Bello}.} \bibinfo{year}{2018}\natexlab{}.
\newblock \showarticletitle{Crepe: A convolutional representation for pitch
  estimation}. In \bibinfo{booktitle}{\emph{2018 IEEE International Conference
  on Acoustics, Speech and Signal Processing (ICASSP)}}. IEEE,
  \bibinfo{publisher}{IEEE}, \bibinfo{address}{New York, NY, USA},
  \bibinfo{pages}{161--165}.
\newblock


\bibitem[Kratky et~al\mbox{.}(2025)]%
        {kratky2025catoracollisionawaretimeoptimalformation}
\bibfield{author}{\bibinfo{person}{Vit Kratky}, \bibinfo{person}{Robert
  Penicka}, \bibinfo{person}{Jiri Horyna}, \bibinfo{person}{Petr Stibinger},
  \bibinfo{person}{Tomas Baca}, \bibinfo{person}{Matej Petrlik},
  \bibinfo{person}{Petr Stepan}, {and} \bibinfo{person}{Martin Saska}.}
  \bibinfo{year}{2025}\natexlab{}.
\newblock \bibinfo{title}{CAT-ORA: Collision-Aware Time-Optimal Formation
  Reshaping for Efficient Robot Coordination in 3D Environments}.
\newblock


\bibitem[Le~Goc et~al\mbox{.}(2016)]%
        {legoc2016zooids}
\bibfield{author}{\bibinfo{person}{Mathieu Le~Goc}, \bibinfo{person}{Lawrence~H
  Kim}, \bibinfo{person}{Ali Parsaei}, \bibinfo{person}{Jean-Daniel Fekete},
  \bibinfo{person}{Pierre Dragicevic}, {and} \bibinfo{person}{Sean Follmer}.}
  \bibinfo{year}{2016}\natexlab{}.
\newblock \showarticletitle{Zooids: Building blocks for swarm user interfaces}.
  In \bibinfo{booktitle}{\emph{Proceedings of the 29th Annual Symposium on User
  Interface Software and Technology}}. \bibinfo{publisher}{ACM},
  \bibinfo{address}{New York, NY, USA}, \bibinfo{pages}{97--109}.
\newblock


\bibitem[McCormack et~al\mbox{.}(2019)]%
        {mccormack2019holon}
\bibfield{author}{\bibinfo{person}{Jon McCormack}, \bibinfo{person}{Toby
  Gifford}, \bibinfo{person}{Patrick Hutchings}, \bibinfo{person}{Maria~Teresa
  Llano}, \bibinfo{person}{Matthew Yee-King}, {and} \bibinfo{person}{Mark
  d'Inverno}.} \bibinfo{year}{2019}\natexlab{}.
\newblock \showarticletitle{Holon: A Framework for Simulation and Exhibition of
  Artificial Biota}. In \bibinfo{booktitle}{\emph{Proceedings of the 10th
  International Conference on Computational Creativity (ICCC)}}.
  \bibinfo{publisher}{Association for Computational Creativity},
  \bibinfo{address}{Charlotte, NC, USA}, \bibinfo{numpages}{8}~pages.
\newblock
\newblock
\shownote{\emph{Holon} (2018) is an interactive aquatic robotic ecosystem.}.


\bibitem[McFee et~al\mbox{.}(2015)]%
        {mcfee2015librosa}
\bibfield{author}{\bibinfo{person}{Brian McFee}, \bibinfo{person}{Colin
  Raffel}, \bibinfo{person}{Dawen Liang}, \bibinfo{person}{Daniel~PW Ellis},
  \bibinfo{person}{Matt McVicar}, \bibinfo{person}{Eric Battenberg}, {and}
  \bibinfo{person}{Oriol Nieto}.} \bibinfo{year}{2015}\natexlab{}.
\newblock \showarticletitle{librosa: Audio and music signal analysis in
  python}. In \bibinfo{booktitle}{\emph{Proceedings of the 14th Python in
  Science Conference (SciPy 2015)}}, Vol.~\bibinfo{volume}{8}.
  \bibinfo{publisher}{SciPy}, \bibinfo{address}{Austin, TX, USA},
  \bibinfo{pages}{18--25}.
\newblock


\bibitem[Morgan et~al\mbox{.}(2014)]%
        {morgan2014mpc}
\bibfield{author}{\bibinfo{person}{Daniel Morgan}, \bibinfo{person}{Soon-Jo
  Chung}, {and} \bibinfo{person}{Fred~Y Hadaegh}.}
  \bibinfo{year}{2014}\natexlab{}.
\newblock \showarticletitle{Model Predictive Control of Swarms of Spacecraft
  Using Sequential Convex Programming}.
\newblock \bibinfo{journal}{\emph{Journal of Guidance, Control, and Dynamics}}
  \bibinfo{volume}{37}, \bibinfo{number}{6} (\bibinfo{year}{2014}),
  \bibinfo{pages}{1725--1740}.
\newblock


\bibitem[Pask(1969)]%
        {pask1969architectural}
\bibfield{author}{\bibinfo{person}{Gordon Pask}.}
  \bibinfo{year}{1969}\natexlab{}.
\newblock \showarticletitle{The Architectural Relevance of Cybernetics}.
\newblock \bibinfo{journal}{\emph{Architectural Design}} \bibinfo{volume}{39},
  \bibinfo{number}{9} (\bibinfo{year}{1969}), \bibinfo{pages}{494--496}.
\newblock


\bibitem[Reynolds(1987)]%
        {reynolds1987flocks}
\bibfield{author}{\bibinfo{person}{Craig~W Reynolds}.}
  \bibinfo{year}{1987}\natexlab{}.
\newblock \showarticletitle{Flocks, Herds, and Schools: A Distributed
  Behavioral Model}. In \bibinfo{booktitle}{\emph{Proceedings of the 14th
  Annual Conference on Computer Graphics and Interactive Techniques (SIGGRAPH
  '87)}}. \bibinfo{publisher}{ACM}, \bibinfo{address}{New York, NY, USA},
  \bibinfo{pages}{25--34}.
\newblock


\bibitem[Rinaldo(2002)]%
        {rinaldo2000autopoiesis}
\bibfield{author}{\bibinfo{person}{Ken Rinaldo}.}
  \bibinfo{year}{2002}\natexlab{}.
\newblock \showarticletitle{Autopoiesis}.
\newblock \bibinfo{journal}{\emph{Leonardo}} \bibinfo{volume}{35},
  \bibinfo{number}{4} (\bibinfo{year}{2002}), \bibinfo{pages}{395--396}.
\newblock
\newblock
\shownote{Artist statement for the robotic sculpture \emph{Autopoiesis}
  (2000).}.


\bibitem[Rogel et~al\mbox{.}(2025)]%
        {weinberg2025dore}
\bibfield{author}{\bibinfo{person}{Amit Rogel}, \bibinfo{person}{Qiaoyu Yang},
  \bibinfo{person}{Jack Hayley}, {and} \bibinfo{person}{Gil Weinberg}.}
  \bibinfo{year}{2025}\natexlab{}.
\newblock \showarticletitle{Do Re Mi Fa So Pass the Tool: Using Melodic
  Prediction to Improve Human-Robot Fluency}. In \bibinfo{booktitle}{\emph{2025
  34th IEEE International Conference on Robot and Human Interactive
  Communication (RO-MAN)}}. IEEE, \bibinfo{publisher}{IEEE},
  \bibinfo{address}{New York, NY, USA}, \bibinfo{pages}{199--206}.
\newblock


\bibitem[Rouard et~al\mbox{.}(2023)]%
        {rouard2023hybrid}
\bibfield{author}{\bibinfo{person}{Simon Rouard}, \bibinfo{person}{Francisco
  Massa}, {and} \bibinfo{person}{Alexandre D{\'e}fossez}.}
  \bibinfo{year}{2023}\natexlab{}.
\newblock \showarticletitle{Hybrid transformers for music source separation}.
  In \bibinfo{booktitle}{\emph{ICASSP 2023-2023 IEEE International Conference
  on Acoustics, Speech and Signal Processing (ICASSP)}}. IEEE,
  \bibinfo{publisher}{IEEE}, \bibinfo{address}{New York, NY, USA},
  \bibinfo{pages}{1--5}.
\newblock


\bibitem[Santos and Egerstedt(2020)]%
        {santos2020interactive}
\bibfield{author}{\bibinfo{person}{Maria Santos} {and} \bibinfo{person}{Magnus
  Egerstedt}.} \bibinfo{year}{2020}\natexlab{}.
\newblock \showarticletitle{Interactive Multi-Robot Painting Through Colored
  Motion Trails}.
\newblock \bibinfo{journal}{\emph{Frontiers in Robotics and AI}}
  \bibinfo{volume}{7} (\bibinfo{year}{2020}), \bibinfo{pages}{580415}.
\newblock


\bibitem[Savery et~al\mbox{.}(2023)]%
        {weinberg2023happy}
\bibfield{author}{\bibinfo{person}{Richard Savery}, \bibinfo{person}{Amit
  Rogel}, \bibinfo{person}{Lisa Zahray}, {and} \bibinfo{person}{Gil Weinberg}.}
  \bibinfo{year}{2023}\natexlab{}.
\newblock \showarticletitle{How Happy Should I be? Leveraging Neuroticism and
  Extraversion for Music-Driven Emotional Interaction in Robotics}.
\newblock In \bibinfo{booktitle}{\emph{Sound and Robotics}}.
  \bibinfo{publisher}{Chapman and Hall/CRC}, \bibinfo{address}{Boca Raton, FL,
  USA}, \bibinfo{pages}{199--218}.
\newblock


\bibitem[Schuck et~al\mbox{.}(2025)]%
        {schuck2025swarmgpt}
\bibfield{author}{\bibinfo{person}{Martin Schuck},
  \bibinfo{person}{Dinushka~Orrin Dahanaggamaarachchi}, \bibinfo{person}{Ben
  Sprenger}, \bibinfo{person}{Vedant Vyas}, \bibinfo{person}{Siqi Zhou}, {and}
  \bibinfo{person}{Angela~P Schoellig}.} \bibinfo{year}{2025}\natexlab{}.
\newblock \showarticletitle{SwarmGPT: Combining Large Language Models with Safe
  Motion Planning for Drone Swarm Choreography}.
\newblock \bibinfo{journal}{\emph{IEEE Robotics and Automation Letters}}
  \bibinfo{volume}{10}, \bibinfo{number}{11} (\bibinfo{year}{2025}),
  \bibinfo{pages}{12237--12244}.
\newblock


\bibitem[Turpin et~al\mbox{.}(2014)]%
        {turpin2014capt}
\bibfield{author}{\bibinfo{person}{Matthew Turpin}, \bibinfo{person}{Nathan
  Michael}, {and} \bibinfo{person}{Vijay Kumar}.}
  \bibinfo{year}{2014}\natexlab{}.
\newblock \showarticletitle{Capt: Concurrent assignment and planning of
  trajectories for multiple robots}.
\newblock \bibinfo{journal}{\emph{The International Journal of Robotics
  Research}} \bibinfo{volume}{33}, \bibinfo{number}{1} (\bibinfo{year}{2014}),
  \bibinfo{pages}{98--112}.
\newblock
\urldef\tempurl%
\url{https://doi.org/10.1177/0278364913515307}
\showDOI{\tempurl}


\bibitem[Virtanen et~al\mbox{.}(2020)]%
        {virtanen2020scipy}
\bibfield{author}{\bibinfo{person}{Pauli Virtanen}, \bibinfo{person}{Ralf
  Gommers}, \bibinfo{person}{Travis~E Oliphant}, \bibinfo{person}{Matt
  Haberland}, \bibinfo{person}{Tyler Reddy}, \bibinfo{person}{David
  Cournapeau}, \bibinfo{person}{Evgeni Burovski}, \bibinfo{person}{Pearu
  Peterson}, \bibinfo{person}{Warren Weckesser}, \bibinfo{person}{Jonathan
  Bright}, \bibinfo{person}{St{\'e}fan~J {van der Walt}},
  \bibinfo{person}{Matthew Brett}, \bibinfo{person}{Joshua Wilson},
  \bibinfo{person}{K~Jarrod Millman}, \bibinfo{person}{Nikolay Mayorov},
  \bibinfo{person}{Andrew R~J Nelson}, \bibinfo{person}{Eric Jones},
  \bibinfo{person}{Robert Kern}, \bibinfo{person}{Eric Larson},
  \bibinfo{person}{C~J Carey}, \bibinfo{person}{{\.I}lhan Polat},
  \bibinfo{person}{Yu Feng}, \bibinfo{person}{Eric~W Moore},
  \bibinfo{person}{Jake {VanderPlas}}, \bibinfo{person}{Denis Laxalde},
  \bibinfo{person}{Josef Perktold}, \bibinfo{person}{Robert Cimrman},
  \bibinfo{person}{Ian Henriksen}, \bibinfo{person}{E~A Quintero},
  \bibinfo{person}{Charles~R Harris}, \bibinfo{person}{Anne~M Archibald},
  \bibinfo{person}{Ant{\^o}nio~H Ribeiro}, \bibinfo{person}{Fabian Pedregosa},
  \bibinfo{person}{Paul {van Mulbregt}}, {and} \bibinfo{person}{{SciPy 1.0
  Contributors}}.} \bibinfo{year}{2020}\natexlab{}.
\newblock \showarticletitle{SciPy 1.0: Fundamental Algorithms for Scientific
  Computing in Python}.
\newblock \bibinfo{journal}{\emph{Nature Methods}}  \bibinfo{volume}{17}
  (\bibinfo{year}{2020}), \bibinfo{pages}{261--272}.
\newblock
\urldef\tempurl%
\url{https://doi.org/10.1038/s41592-019-0686-2}
\showDOI{\tempurl}


\bibitem[Vorn(2016)]%
        {vorn2017hysterical}
\bibfield{author}{\bibinfo{person}{Bill Vorn}.}
  \bibinfo{year}{2016}\natexlab{}.
\newblock \showarticletitle{Artistic Approaches in Robotics: A Personal
  Journey}. In \bibinfo{booktitle}{\emph{Robotic Art: Exploring an Unlikely
  Symbiosis}}, \bibfield{editor}{\bibinfo{person}{Damith Herath},
  \bibinfo{person}{Christian Kroos}, {and} \bibinfo{person}{Stelarc}} (Eds.).
  \bibinfo{publisher}{Springer}, \bibinfo{address}{Cham, Switzerland},
  \bibinfo{pages}{243--262}.
\newblock
\newblock
\shownote{Discusses works including \emph{Hysterical Machines} and \emph{Grace
  State Machines}.}.


\bibitem[Waibel et~al\mbox{.}(2017)]%
        {waibel2017drone}
\bibfield{author}{\bibinfo{person}{Markus Waibel}, \bibinfo{person}{Bill
  Keays}, {and} \bibinfo{person}{Federico Augugliaro}.}
  \bibinfo{year}{2017}\natexlab{}.
\newblock \bibinfo{title}{Drone shows: Creative potential and best practices}.
\newblock \bibinfo{howpublished}{Verity Studios White Paper}.
\newblock
\urldef\tempurl%
\url{https://veritystudios.com}
\showURL{%
\tempurl}


\bibitem[Wang et~al\mbox{.}(2020)]%
        {wang2020roboat}
\bibfield{author}{\bibinfo{person}{Wei Wang}, \bibinfo{person}{Luis~A Mateos},
  \bibinfo{person}{Shuguang Park}, \bibinfo{person}{Pietro Galdems},
  \bibinfo{person}{Dylan Kellogg}, \bibinfo{person}{Fabio Duarte},
  \bibinfo{person}{Carlo Ratti}, {and} \bibinfo{person}{Daniela Rus}.}
  \bibinfo{year}{2020}\natexlab{}.
\newblock \showarticletitle{Roboat II: A novel autonomous surface vessel for
  urban environments}. In \bibinfo{booktitle}{\emph{2020 IEEE/RSJ International
  Conference on Intelligent Robots and Systems (IROS)}}. IEEE,
  \bibinfo{publisher}{IEEE}, \bibinfo{address}{New York, NY, USA},
  \bibinfo{pages}{1740--1747}.
\newblock


\bibitem[Weinberg and Driscoll(2006)]%
        {weinberg2006toward}
\bibfield{author}{\bibinfo{person}{Gil Weinberg} {and} \bibinfo{person}{Scott
  Driscoll}.} \bibinfo{year}{2006}\natexlab{}.
\newblock \showarticletitle{Toward robotic musicianship}.
\newblock \bibinfo{journal}{\emph{Computer Music Journal}}
  \bibinfo{volume}{30}, \bibinfo{number}{4} (\bibinfo{year}{2006}),
  \bibinfo{pages}{28--45}.
\newblock


\bibitem[Wicaksono et~al\mbox{.}(2024)]%
        {wicaksono2024knitwork}
\bibfield{author}{\bibinfo{person}{Irma Wicaksono}, \bibinfo{person}{Lancelot
  Blanchard}, \bibinfo{person}{Sam Chin}, \bibinfo{person}{Cristian Colon},
  {and} \bibinfo{person}{Joseph~A Paradiso}.} \bibinfo{year}{2024}\natexlab{}.
\newblock \showarticletitle{KnitworkVR: Dual-reality Experience through
  Distributed Sensor-Actuator Networks in the Living Knitwork Pavilion}. In
  \bibinfo{booktitle}{\emph{SIGGRAPH Asia 2024 Emerging Technologies}}.
  \bibinfo{publisher}{ACM}, \bibinfo{address}{New York, NY, USA},
  \bibinfo{numpages}{2}~pages.
\newblock


\bibitem[Wicaksono et~al\mbox{.}(2022)]%
        {wicaksono2022tapis}
\bibfield{author}{\bibinfo{person}{Irma Wicaksono}, \bibinfo{person}{Don~Derek
  Haddad}, {and} \bibinfo{person}{Joseph~A Paradiso}.}
  \bibinfo{year}{2022}\natexlab{}.
\newblock \showarticletitle{Tapis Magique: Machine-knitted Electronic Textile
  Carpet for Interactive Choreomusical Performance and Immersive Environments}.
  In \bibinfo{booktitle}{\emph{Proceedings of the 2022 ACM Conference on
  Creativity and Cognition}}. \bibinfo{publisher}{ACM}, \bibinfo{address}{New
  York, NY, USA}, \bibinfo{pages}{326--339}.
\newblock
\urldef\tempurl%
\url{https://doi.org/10.1145/3527927.3532812}
\showDOI{\tempurl}


\end{thebibliography}

\end{document}